\pdfoutput=1
\documentclass{article}

\usepackage[preprint]{neurips_2025}

\usepackage[utf8]{inputenc}
\usepackage[T1]{fontenc}
\usepackage{hyperref}
\usepackage{url}
\usepackage{booktabs}
\usepackage{amsfonts}
\usepackage{amsmath}
\usepackage{microtype}
\usepackage{xcolor}
\usepackage{graphicx}

\newcommand{\tabpfn}{TabPFN-3.5}
\newcommand{\deepl}{DeepED\textsubscript{LSTM}}
\newcommand{\deept}{DeepED\textsubscript{Transformer}}

\title{Tabular Foundation Models as Emulators for Forecasting Terrestrial Ecosystem Dynamics}

\author{%
  Zhihao Wang$^1$, \quad Ruohan Li$^1$, \quad Lei Ma$^1$, \quad George Hurtt$^1$, \quad Xiaowei Jia$^2$, \quad Yiqun Xie$^1$\thanks{Corresponding author.}\\
  $^1$University of Maryland, \quad $^2$Rutgers University\\
  \texttt{\{zhwang1, r526li, lma6, gchurtt, xie\}@umd.edu,}
  \texttt{xiaowei.jia@rutgers.edu}
}
\begin{document}

\maketitle

\begin{abstract}
Process-based terrestrial biosphere models such as the Ecosystem Demography (ED) model underpin global carbon budgets, but
their cost limits global, high-resolution and scenario-rich simulation. Deep emulators such as DeepED reduce that cost but
require task-specific architectures and training. Tabular foundation models offer a different route: TabPFN predicts new
rows by in-context learning, without being trained on the task. We compare \tabpfn{} with two DeepED baselines, \deepl{}
and \deept{}, on CarbonGlobe's overall evaluation, which trains on a uniform 1/16 sample of sites and tests on the other
50,779, predicting the 40-year state of all seven ED outputs. We then test every model, trained on present-day data only,
on ED runs forced by SSP1-2.6 and SSP5-8.5 for 2021--2060 in six continental forest regions. With no training, \tabpfn{}
has the lowest present-day height error of the three models and is within 4\% of \deepl{} for soil carbon, but is less
accurate for AGB, LAI and the carbon fluxes. In 2060, errors are several times larger than at the same sites under
present-day forcing, and every model underestimates the biomass that ED gains. Under SSP5-8.5, \tabpfn{} is the most
accurate model for height, AGB, soil carbon and LAI, with the lowest AGB error in all six regions, although, unlike ED, it
predicts no larger biomass gain than under SSP1-2.6. Under SSP1-2.6, no model is best overall.
\end{abstract}

\section{Introduction}

Forests store a large share of terrestrial carbon, and how much they keep absorbing under a changing climate is a central
uncertainty of the global carbon cycle~\citep{friedlingstein2023gcb}. Process-based models answer such questions by
simulating vegetation demography and biogeochemistry explicitly. The Ecosystem Demography model (ED;
\citealp{moorcroft2001ed,hurtt1998terrestrial,ma2022edv3}) represents the size- and age-structure of forest stands and their
carbon fluxes; it contributes to the Global Carbon Budget and supports NASA's Carbon Monitoring
System~\citep{wang2025carbonglobe}. Its cost, however, grows with spatial resolution, the number of scenarios and the length
of the simulation, which limits exactly the global, scenario-rich runs that policy needs.

Machine-learning emulators trained on ED outputs can replace most of this computation. DeepED~\citep{wang2023deeped}
emulates ED's 40-year carbon trajectories with recurrent networks and age-aware inputs, and the CarbonGlobe
benchmark~\citep{wang2025carbonglobe} made such emulation a standard machine-learning task: global coverage at
0.5$^\circ$, 54,152 sites, 136 monthly drivers, 15 initial forest ages and seven carbon and vegetation outputs over 40
years. These emulators are effective, but each is a task-specific architecture that must be designed, trained and tuned.

Foundation models suggest another route. Pretrained once at scale, they are applied to new tasks with little or no
task-specific training~\citep{bommasani2021opportunities}; in Earth science, foundation models now forecast weather and
climate~\citep{nguyen2023climax,bodnar2025aurora}, and time-series foundation models forecast unseen series
zero-shot~\citep{ansari2024chronos,das2024timesfm}. For tabular data, TabPFN~\citep{hollmann2023tabpfn,hollmann2025tabpfnv2}
is a transformer pretrained on synthetic tasks that predicts the targets of new rows by in-context learning: the training
rows are its context, and no weights are updated. Its newest version, \tabpfn{}~\citep{jaeger2026tabpfn35}, takes contexts
far larger than the ten thousand rows that TabPFN~v2 was built for. Whether such a model can emulate a process-based
ecosystem model, and how it compares with purpose-built emulators in the present and in a future climate, has not been
tested.

We explore this question on CarbonGlobe. We cast ED emulation as tabular regression, with one row per forest stand (a site
started from one of its initial states) and one in-context fit per output variable, and compare \tabpfn{} with two DeepED
baselines: \deepl{}, built on a recurrent network, and \deept{}, built on a Transformer. Our contributions are:
\begin{itemize}
  \item \textbf{A tabular foundation model as an ED emulator.} Without any training, \tabpfn{} predicts the 40-year state
    of all seven ED outputs from in-context fits on CarbonGlobe's training sites (Section~\ref{sec:methods}).
  \item \textbf{A present-day comparison with trained emulators.} On CarbonGlobe's overall evaluation, 50,779 test sites,
    \tabpfn{} has the lowest height error and is close to the best baseline for soil carbon, but is less accurate for
    AGB, LAI and the carbon fluxes (Section~\ref{sec:present}).
  \item \textbf{A future-climate test in six continental forest regions.} All models, trained on present-day data only,
    are evaluated on ED runs under SSP1-2.6 and SSP5-8.5 for 2021--2060 at 1,858 forest sites in six regions, one per
    continent, and compared with the same sites under present-day forcing (Figure~\ref{fig:study};
    Section~\ref{sec:future}). Errors grow several-fold, and every model underestimates the biomass ED gains by 2060.
    Under SSP5-8.5, \tabpfn{} is the most accurate model for height, AGB, soil carbon and LAI.
\end{itemize}

\section{Related work}

\paragraph{Process-based models and their emulators.} Vegetation demographic models such as ED represent the size and age
structure of forests~\citep{moorcroft2001ed,fisher2018vegetation}. Machine learning increasingly emulates or complements
process models in Earth science~\citep{reichstein2019deep}, often guided by domain knowledge~\citep{karpatne2017theory,
jia2021physics}. Emulators of land models support parameter estimation~\citep{dagon2020clm5}, and deep emulators now
approximate numerical weather prediction at global scale~\citep{kurth2023fourcastnet,lam2023graphcast,bi2023pangu}, with
benchmarks such as WeatherBench~2~\citep{rasp2023weatherbench2}. For forest carbon, DeepED~\citep{wang2023deeped}
introduced long-horizon emulation of ED with age-aware recurrent networks, and CarbonGlobe~\citep{wang2025carbonglobe}
released a global ML-ready dataset with recurrent, transformer-based~\citep{vaswani2017attention,zhou2021informer} and
linear baselines, evaluated with the RMSE, the MAE and two problem-driven metrics, the delta and cumulative errors.
EcoDiffusion~\citep{li2026ecodiffusion} makes such emulation uncertainty-aware, generating long sequences from a single
initial step with a conditional diffusion model.

\paragraph{Emulators under climate change.} What-if simulations, such as future climate scenarios, are a central use of
emulators, and an Earth system world model has recently been built for them, with terrestrial ecosystems as its case
study~\citep{wang2026earth}. An emulator trained on the present climate, however, has to extrapolate in a future scenario:
temperature, precipitation and CO$_2$ leave the range it was trained on. \citet{beucler2024climate} showed that
machine-learning parameterizations trained in one climate generalize poorly to a warmer one unless their inputs are made
climate-invariant. We test this setting for ED emulators, with scenario runs at sites never used for training.

\paragraph{Foundation models for science and time series.} Pretrained models for the atmosphere and Earth system
\citep{nguyen2023climax,bodnar2025aurora} and for time series \citep{ansari2024chronos,das2024timesfm,woo2024moirai}
transfer to new tasks with little training. Time-series models forecast from past observations of the series itself; ED
emulation instead starts from an initial ecosystem state and a long sequence of forcing, a setting closer to conditional
regression.

\paragraph{Tabular foundation models.} TabPFN~\citep{hollmann2023tabpfn} showed that a transformer pretrained on synthetic
data can solve small tabular tasks in context; TabPFN~v2~\citep{hollmann2025tabpfnv2} extended this to regression and to
datasets of up to about ten thousand rows. Later models scale the context further~\citep{qu2025tabicl,ma2024tabdpt,
jaeger2026tabpfn35} or retrieve a local context for each test row~\citep{thomas2024retrieval}. To our knowledge,
tabular foundation models have not been evaluated as emulators of process-based ecosystem models, nor under a climate
shift.

\section{Problem definition}
\label{sec:problem}

\paragraph{Data.} CarbonGlobe~\citep{wang2025carbonglobe} contains ED v3.0 simulations at 54,152 land sites on a
0.5$^\circ$ global grid. At each site, ED is started from 15 initial forest states, taken at stand ages
$a \in \{1, 10, 20, 30, 50, 70, 100, 150, 200, 250, 300, 350, 400, 450, 500\}$ years, and run for the 40 years from 1981
under observation-based forcing (MERRA-2 meteorology and NOAA CarbonTracker CO$_2$). We call a site started from one of its
initial states a \emph{stand}. The forcing $X \in \mathbb{R}^{40 \times 12 \times 136}$ holds 136 monthly drivers: air
temperature, precipitation and soil temperature at six depths (8); disturbance (1); diurnal cycles, as 24 hourly values, of
atmospheric CO$_2$, specific humidity, air temperature, shortwave radiation and wind speed (120); and seven soil properties.
The state $y_t \in \mathbb{R}^7$ is annual: vegetation height (m), aboveground biomass (AGB), soil carbon, leaf area index
(LAI, m$^2$\,m$^{-2}$), gross and net primary production (GPP, NPP) and heterotrophic respiration (Rh); carbon stocks and
fluxes are in kg\,C\,m$^{-2}$ (fluxes as annual totals).

\paragraph{Task.} Given a stand's initial state $y_0$ at age $a$ and the 40-year forcing $X$ of its site, predict the
state of all seven outputs after 40 years, $y_{40}$. The deep emulators produce the whole trajectory $y_1, \dots, y_{40}$
on the way; \tabpfn{} predicts $y_{40}$ directly.

\paragraph{Present-day evaluation.} We follow CarbonGlobe's overall evaluation~\citep{wang2025carbonglobe}. Models learn
from a uniform 1/16 sample of the sites, every fourth cell in latitude and longitude (\emph{train4}: 3,373 sites on a
2$^\circ$ grid, of which 337 are held out for early stopping), and are tested on all other 50,779 sites (761,685 stands).
Uniform sampling mirrors how an emulator would be used: ED runs on a coarse grid train it, and it then fills in the fine
grid.

\paragraph{Future-climate runs.} \label{sec:future-def} To test generalization beyond the present-day climate, we use ED
runs forced by the GFDL-ESM4 Earth system model~\citep{dunne2020gfdlesm4} under SSP1-2.6 and SSP5-8.5, a low- and a
high-emission scenario~\citep{oneill2016scenariomip}. They start in 2020 from 15 initial states per site, at the same stand
ages as above except that the youngest is 0 years, and we use their first 40 years, 2021--2060. These runs differ from the
present-day runs in two ways: the climate changes, and the forcing comes from a climate model instead of observations.

\begin{figure}[t]
  \centering
  \includegraphics[width=\linewidth]{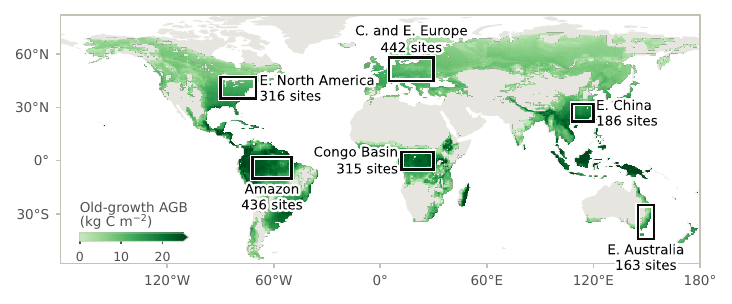}
  \caption{Study area of the future-climate evaluation. Shading: aboveground biomass (AGB) of the oldest initial state
    (500-year stand) of each CarbonGlobe cell where that stand is at least 5\,m tall; other land in gray. Boxes: the six
    regions, with the number of sites used in each (Section~\ref{sec:future-def}, Table~\ref{tab:regions}).}
  \label{fig:study}
\end{figure}

\paragraph{Six forest regions.} We evaluate on six forest regions, one per continent, that span tropical moist,
subtropical, temperate and eucalypt forest (Figure~\ref{fig:study}, Table~\ref{tab:regions}). Every site we use is a
0.5$^\circ$ cell inside a region that (i) is not in CarbonGlobe's training or validation split and (ii) is forest: its oldest
initial state, a 500-year stand, is at least 5\,m tall in 2020 under both scenarios, the tree-height threshold of the FAO
forest definition~\citep{fao2018fraterms}. In all, we use 1,858 sites, each with 15 initial states. All models learn from present-day train4 data only, so they have
never seen these locations. Over 2021--2060 the regions are 0.9--1.2\,K (SSP1-2.6) and 1.2--1.8\,K (SSP5-8.5) warmer than
in their present-day forcing. Atmospheric CO$_2$, 411--442\,ppm in the last present-day year at these sites, reaches
474\,ppm (SSP1-2.6) and 643\,ppm (SSP5-8.5) by 2060.

\begin{table}[t]
  \caption{The six future-climate regions. Sites: 0.5$^\circ$ cells used, each with 15 initial states. AGB: median AGB of the
    500-year stand in 2020 (kg\,C\,m$^{-2}$). $T$: mean annual air temperature of the present-day forcing ($^\circ$C).
    $\Delta T$: mean temperature of the GFDL-ESM4 forcing over 2021--2060 minus $T$ (K); it includes the difference between
    the climate model and the observation-based present-day forcing.}
  \label{tab:regions}
  \centering
  \small
  \setlength{\tabcolsep}{3.5pt}
  \begin{tabular}{lllrrrrr}
    \toprule
    & & & & & & \multicolumn{2}{c}{$\Delta T$} \\
    \cmidrule(l){7-8}
    Region & Bounds & Forest & Sites & AGB & $T$ & SSP1-2.6 & SSP5-8.5 \\
    \midrule
    E. North America & 35--47$^\circ$N, 90--70$^\circ$W & temperate broadleaf & 316 & 12.2 & 9.1 & +1.2 & +1.8 \\
    Amazon & 10$^\circ$S--2$^\circ$N, 72--50$^\circ$W & tropical moist & 436 & 22.2 & 25.7 & +1.1 & +1.5 \\
    C. and E. Europe & 45--58$^\circ$N, 5--30$^\circ$E & temperate mixed & 442 & 10.2 & 8.2 & +1.1 & +1.5 \\
    Congo Basin & 5$^\circ$S--5$^\circ$N, 12--30$^\circ$E & tropical moist & 315 & 20.8 & 24.3 & +0.9 & +1.3 \\
    E. China & 22--32$^\circ$N, 108--120$^\circ$E & subtropical broadleaf & 186 & 18.1 & 17.3 & +1.1 & +1.2 \\
    E. Australia & 44--25$^\circ$S, 145--154$^\circ$E & eucalypt & 163 & 6.3 & 17.3 & +0.9 & +1.2 \\
    \midrule
    All six & & & 1,858 & & & & \\
    \bottomrule
  \end{tabular}
\end{table}

\paragraph{Metric.} For each output variable we report the root-mean-square error at year 40 over all stands,
$\big[\tfrac{1}{N}\sum_i (\hat y_{i,40} - y_{i,40})^2\big]^{1/2}$, in the variable's physical units. This is CarbonGlobe's
cumulative error~\citep{wang2025carbonglobe} in root-mean-square form: the error left at the end of the 40-year forecast,
after it has accumulated over all years.

\section{Methods}
\label{sec:methods}

\subsection{\tabpfn{} as an in-context emulator}
Each stand is one row. Its features are the 40-year mean of each of the 136 drivers, the initial state $y_0$ (7), the
stand age divided by 500 (1), and two age references from DeepED~\citep{wang2023deeped}: the mean initial state of all
sites at that age and its slope across ages (14), 158 features in all. Drivers and states are standardized with
CarbonGlobe's statistics. One regressor predicts the year-40 value of each output variable, seven in all. A fit only
places the context rows, the training stands with that target, in the model's key-value cache: no weights are updated and
nothing is tuned. We use \tabpfn{}~\citep{jaeger2026tabpfn35} with two estimators, one per target transform of the
checkpoint, take the mean of its predictive distribution, and use the 3,036 training sites of train4 as context (45,540
rows). The weights are used under Prior Labs' TabPFN-3.5 Non-Commercial License (v1.0).

\subsection{DeepED baselines}
DeepED~\citep{wang2023deeped} emulates ED with a sequence model that steps from the initial state through the 40 years of
forcing, with the stand age and age-dependent reference states as additional inputs. We use two versions as baselines.

\paragraph{\deepl{}.} We use DeepED's LSTM with an improved design. Each year, an LSTM with 160 units reads the year's 12
monthly drivers together with the previous state, the stand age and the reference states, and a small multilayer
perceptron turns the year's summary into the change of all seven states. The network is trained through the whole 40-year
rollout on its own predictions, so training sees the same error accumulation as forecasting (AdamW with learning rate
$10^{-3}$ halved on plateaus, batches of 1,024 stands, gradient clipping at 1, early stopping on the held-out sites). It
has 0.29\,M parameters.

\paragraph{\deept{}.} An Informer encoder--decoder~\citep{zhou2021informer} replaces the recurrent network and predicts all
40 years in one pass. Each token is one month. The encoder reads 12 tokens, the first year's drivers with the initial
state; the decoder reads these 12 tokens followed by the 480 forecast months with the state set to zero, and its output at
the last month of each forecast year is that year's state. The tokens also carry the stand age and the reference states.
It has 11.8\,M parameters and is trained on the squared error of the states plus half the squared error of their
year-to-year changes (Adam with learning rate $10^{-4}$ halved every epoch, batches of 16 stands, early stopping on the
held-out sites).

\subsection{Protocol}
Each deep baseline is trained with three seeds, and we report the mean prediction of the three. Early stopping uses only
the 337 held-out train4 sites, and the six regions were fixed before any model was scored on them. Future inputs are
standardized with the same present-day statistics as the training data. All experiments run on NVIDIA A100 GPUs (40\,GB).

\section{Experiments}

\subsection{Present day}
\label{sec:present}
Table~\ref{tab:v1present} gives the year-40 RMSE of each output on CarbonGlobe's 50,779 test sites.

\begin{table}[t]
  \caption{Present day, CarbonGlobe's overall evaluation: year-40 RMSE over the 50,779 test sites (761,685 stands); every
    model learns from train4 only. Height in m, LAI in m$^2$\,m$^{-2}$, the other variables in kg\,C\,m$^{-2}$ (fluxes
    per year). \deepl{} and \deept{}: mean prediction of three training seeds. Best per column in bold.}
  \label{tab:v1present}
  \centering
  \small
  \begin{tabular}{lrrrrrrr}
    \toprule
    Model & Height & AGB & Soil C & LAI & GPP & NPP & Rh \\
    \midrule
    \quad \tabpfn{} & \textbf{1.617} & 0.585 & 0.551 & 0.377 & 0.256 & 0.127 & 0.143 \\
    \quad \deepl{} & 1.667 & \textbf{0.473} & \textbf{0.533} & \textbf{0.306} & \textbf{0.172} & \textbf{0.085} & \textbf{0.102} \\
    \quad \deept{} & 1.856 & 0.631 & 0.660 & 0.353 & 0.204 & 0.101 & 0.125 \\
    \bottomrule
  \end{tabular}
\end{table}

\paragraph{Without training, \tabpfn{} is competitive for height and soil carbon.} \deepl{} is the most accurate of the
three models for six of the seven outputs (Table~\ref{tab:v1present}). \tabpfn{}, which is never trained, has the lowest
height error of the three (1.62\,m, against 1.67\,m for \deepl{} and 1.86\,m for \deept{}), and its soil-carbon error is
within 4\% of \deepl{}'s. Its errors are larger for AGB (+24\%) and LAI (+23\%), and largest for the three fluxes (+40 to
+49\%). A likely reason is that annual fluxes follow the weather of the final year, which the 40-year mean drivers seen by
\tabpfn{} do not resolve. \deept{} is less accurate than \deepl{} for every output, and less accurate than \tabpfn{} for
height, AGB and soil carbon. Averaging three training seeds helps the deep baselines: single \deepl{} seeds have AGB errors
of 0.515--0.529\,kg\,C\,m$^{-2}$, their mean prediction 0.473.

\subsection{Future climate in six regions}
\label{sec:future}
Table~\ref{tab:v1future} compares each model's year-40 error at the 1,858 region sites under present-day forcing
with its error in 2060 under the two scenarios, and Table~\ref{tab:v1regions} breaks the AGB error down by region.

\begin{table}[t]
  \caption{Six future regions, 1,858 sites (27,870 stands): year-40 RMSE under present-day forcing (the same sites inside
    CarbonGlobe's test set) and in 2060 under SSP1-2.6 and SSP5-8.5. All models learn from present-day train4 data only.
    Units as in Table~\ref{tab:v1present}; best per column and block in bold.}
  \label{tab:v1future}
  \centering
  \small
  \begin{tabular}{lrrrrrrr}
    \toprule
    Model & Height & AGB & Soil C & LAI & GPP & NPP & Rh \\
    \midrule
    \multicolumn{8}{l}{\textit{Present day, the same sites}} \\
    \quad \tabpfn{} & \textbf{1.061} & 0.688 & 0.517 & 0.454 & 0.273 & 0.134 & 0.176 \\
    \quad \deepl{} & 1.233 & \textbf{0.528} & \textbf{0.462} & \textbf{0.311} & \textbf{0.180} & \textbf{0.088} & \textbf{0.159} \\
    \quad \deept{} & 1.412 & 0.909 & 0.683 & 0.413 & 0.230 & 0.114 & 0.181 \\
    \multicolumn{8}{l}{\textit{SSP1-2.6, 2060}} \\
    \quad \tabpfn{} & 2.809 & \textbf{2.066} & 1.482 & 1.461 & 1.146 & 0.582 & 0.457 \\
    \quad \deepl{} & 3.749 & 4.225 & \textbf{1.432} & \textbf{0.836} & \textbf{0.684} & \textbf{0.351} & \textbf{0.247} \\
    \quad \deept{} & \textbf{2.544} & 2.407 & 1.621 & 1.321 & 0.884 & 0.467 & 0.402 \\
    \multicolumn{8}{l}{\textit{SSP5-8.5, 2060}} \\
    \quad \tabpfn{} & \textbf{3.565} & \textbf{2.746} & \textbf{1.966} & \textbf{1.243} & 1.464 & 0.737 & 0.513 \\
    \quad \deepl{} & 6.154 & 6.510 & 2.170 & 1.780 & \textbf{1.006} & \textbf{0.503} & \textbf{0.462} \\
    \quad \deept{} & 6.539 & 6.212 & 3.661 & 1.436 & 1.672 & 0.857 & 0.493 \\
    \bottomrule
  \end{tabular}
\end{table}

\begin{table}[t]
  \caption{Year-40 AGB RMSE (kg\,C\,m$^{-2}$) in each region (Table~\ref{tab:regions}): present-day forcing at the same
    sites, and 2060 under SSP1-2.6 and SSP5-8.5. Best per column and block in bold.}
  \label{tab:v1regions}
  \centering
  \small
  \begin{tabular}{lrrrrrr}
    \toprule
    & E.\ North & & C.\ and E. & Congo & & E. \\
    Model & America & Amazon & Europe & Basin & E.\ China & Australia \\
    \midrule
    \multicolumn{7}{l}{\textit{Present day, the same sites}} \\
    \quad \tabpfn{} & \textbf{0.20} & 1.17 & \textbf{0.33} & 0.61 & \textbf{0.29} & 0.73 \\
    \quad \deepl{} & 0.21 & \textbf{0.82} & 0.35 & \textbf{0.48} & 0.35 & \textbf{0.63} \\
    \quad \deept{} & 0.29 & 1.54 & 0.43 & 0.77 & 0.47 & 1.00 \\
    \multicolumn{7}{l}{\textit{SSP1-2.6, 2060}} \\
    \quad \tabpfn{} & 1.26 & \textbf{2.36} & \textbf{1.04} & 2.96 & 2.49 & 1.93 \\
    \quad \deepl{} & \textbf{0.60} & 8.31 & 1.20 & \textbf{1.79} & \textbf{2.20} & \textbf{1.50} \\
    \quad \deept{} & 1.44 & 2.75 & 1.53 & 3.51 & 2.60 & 2.00 \\
    \multicolumn{7}{l}{\textit{SSP5-8.5, 2060}} \\
    \quad \tabpfn{} & \textbf{2.03} & \textbf{2.41} & \textbf{1.61} & \textbf{4.12} & \textbf{4.04} & \textbf{2.03} \\
    \quad \deepl{} & 3.77 & 10.27 & 4.01 & 5.61 & 6.91 & 3.81 \\
    \quad \deept{} & 4.25 & 6.84 & 3.64 & 9.13 & 7.66 & 3.95 \\
    \bottomrule
  \end{tabular}
\end{table}

\paragraph{Errors grow several-fold, and every model underestimates ED's biomass gain.} Under present-day forcing, the
errors at the region sites are of the same order as on the whole test set (Table~\ref{tab:v1future}). In 2060 the same
models are much less accurate: height and AGB errors are 2--8 times larger under SSP1-2.6 and 3--12 times larger under
SSP5-8.5. Averaged over the region stands, ED's AGB grows between 2020 and 2060 by 1.4\,kg\,C\,m$^{-2}$ under SSP1-2.6
and by 2.3 under SSP5-8.5, to 11.1 and 12.1. \tabpfn{} predicts smaller gains, 0.6 and 0.4, and both DeepED baselines
predict losses: 0.9 and 3.2 for \deepl{}, 0.4 and 2.7 for \deept{}. No model reproduces ED's larger gain under the
high-emission scenario: \tabpfn{} predicts a slightly smaller gain under SSP5-8.5, and the deep baselines larger losses.

\paragraph{Under SSP5-8.5, \tabpfn{} is the most accurate model for height, AGB, soil carbon and LAI.} It also has the
lowest AGB error in all six regions (Table~\ref{tab:v1regions}); over all regions its AGB error, 2.7\,kg\,C\,m$^{-2}$, is
less than half that of \deepl{} (6.5) or \deept{} (6.2). \deepl{} remains the most accurate model for the three fluxes.

\paragraph{Under SSP1-2.6, no model is best overall.} \deepl{} has the lowest errors for soil carbon, LAI and the three
fluxes, and the lowest AGB error in four of the six regions, but its AGB error in the Amazon, 8.3\,kg\,C\,m$^{-2}$ against
2.4 for \tabpfn{}, makes it the least accurate model for AGB over all regions. \tabpfn{} has the lowest AGB error over all
regions, and \deept{} the lowest height error.

\paragraph{The deep baselines fail most in tropical forest, and depend on the training seed.} Their largest AGB errors are
in the two tropical regions: the Amazon for \deepl{} and the Congo Basin for \deept{} (Table~\ref{tab:v1regions}). In the
Amazon, the warmest region, precipitation also falls by 9\% (SSP1-2.6) and 14\% (SSP5-8.5); there \deepl{} predicts AGB to
drop to 5.6\,kg\,C\,m$^{-2}$ by 2060 under SSP5-8.5, while ED keeps it near 15. Their future errors also vary far more
with the training seed than in the present day: under SSP5-8.5, single \deepl{} seeds have AGB errors from 3.7 to
9.6\,kg\,C\,m$^{-2}$ and single \deept{} seeds from 5.8 to 6.9, against 0.515--0.529 and 0.637--0.653 in the present day.

\section{Discussion and limitations}
\paragraph{What the tabular route offers.} Given only CarbonGlobe's 1/16 training sample as context, with no training and
no tuning, a pretrained tabular model predicts ED's 40-year height and soil carbon about as accurately as emulators
designed and trained for the task, though not AGB, LAI and the fluxes. Under the larger climate shift of SSP5-8.5 it is the
most accurate of the three models for height, AGB, soil carbon and LAI, while the trained emulators drift far from ED in
some regions and vary with the training seed.

\paragraph{What no model captures.} \tabpfn{}'s advantage in the future appears to come from responding little to the
change in forcing, not from following ED: its mean 2060 biomass is about the same under both scenarios, whereas ED's is
higher under SSP5-8.5. No model trained on present-day data reproduces the biomass that ED gains by 2060. Emulators meant
for scenario studies therefore need either training data from future climates or knowledge of the processes that make
biomass respond to them.

\paragraph{Limitations.} (i) We compare the models only at year 40. \tabpfn{} can predict any forecast year with its own
in-context fit, which would allow a comparison along the trajectory with CarbonGlobe's delta error. (ii) \tabpfn{}'s
climate features are 40-year means of each driver, so it sees neither the order of events within the 40 years nor the
weather of the final year. (iii) All models learn from train4 only, and \tabpfn{} is not fine-tuned; whether more
present-day data or fine-tuning helps in a future climate is not tested. (iv) The future test uses one Earth system model
and two scenarios, and it changes the source of the forcing as well as the climate.

\section{Conclusion}
We explored a tabular foundation model as an emulator of the ED model on CarbonGlobe. With no training, \tabpfn{} predicts
the 40-year height more accurately than both DeepED baselines and soil carbon within 4\% of the better one, \deepl{}, but
is less accurate for AGB, LAI and the carbon fluxes. In six continental forest regions under SSP1-2.6 and SSP5-8.5, the
errors of every model trained on present-day data grow several-fold, and none reproduces the biomass that ED gains by 2060.
Under SSP5-8.5, \tabpfn{} is the most accurate model for height, AGB, soil carbon and LAI, while the deep baselines' errors
are largest in tropical forest and vary strongly with the training seed.


\bibliographystyle{plainnat}
\bibliography{references}

\end{document}